\documentclass{article} 
\usepackage{iclr2023_conference,times}
\usepackage{helvet}
\usepackage{courier}
\usepackage{xcolor}
\usepackage{siunitx}
\usepackage{graphicx}
\usepackage{amsfonts}
\usepackage{amsmath}
\usepackage{multirow}
\usepackage{float}

\usepackage{amsmath,amsfonts,bm}

\def\eqref#1{equation~\ref{#1}}

\def\1{\bm{1}}

\DeclareMathAlphabet{\mathsfit}{\encodingdefault}{\sfdefault}{m}{sl}
\SetMathAlphabet{\mathsfit}{bold}{\encodingdefault}{\sfdefault}{bx}{n}

\usepackage{hyperref}
\usepackage{url}

\title{Attention-based Domain Adaptation Forecasting of Streamflow in Data-Sparse Regions}

\author{Roland Oruche \\
Department of Electrical Engineering \& Computer Science\\
University of Missouri-Columbia\\
Columbia, MO, 65201, USA \\
\texttt{rro2q2@umsystem.edu} \\
\And
Fearghal O'Donncha \\
IBM Research Europe \\
Dublin, Ireland \\
\texttt{feardonn@ie.ibm.com} \\
}

\newcommand{\Lagr}{\mathcal{L}}
\newcommand{\LagrD}{\mathcal{D}}

\iclrfinalcopy 
\begin{document}

\maketitle

\begin{abstract}
Streamflow forecasts are critical to guide water resource management, mitigate drought and flood effects, and develop climate-smart infrastructure and governance. Many global regions, however, have limited streamflow observations to guide evidence-based management strategies. In this paper, we propose an attention-based domain adaptation streamflow forecaster for data-sparse regions. Our approach leverages the hydrological characteristics of a data-rich source domain to induce effective 24hr lead-time streamflow prediction in a data-constrained target domain. Specifically, we employ a deep-learning framework leveraging domain adaptation techniques to simultaneously train streamflow predictions and discern between both domains using an adversarial method. Experiments against baseline cross-domain forecasting models show improved performance for 24hr lead-time streamflow forecasting.
\end{abstract}


\section{Introduction}
 
Accurate streamflow forecasts are critical to enable sustainable, climate-smart water resource management strategies~\citep{makkeasorn2008short}. Traditionally, engineers have depended on physics-based models, which describe hydrological phenomena using a series of partial differential equations restricted by empirical relationships, heuristics, and specialized knowledge. Although they provide a deep understanding of how water moves over land with respect to time and space, the intricate nature and uncertainty of these models pose a significant challenge for experienced users. Alternatively, machine learning (ML) has emerged as a core tool to develop prediction models that informs effective water resource planning~\citep{kratzert2019toward,nearing2020deep}. However, these models are data hungry and favor regions with well developed environmental monitoring programs providing plentiful observation data to train on.

Transfer learning has shown potential to generalize streamflow forecasts to data-sparse regions. Previous work has applied transfer learning techniques using pre-trained models from data-rich regions to data-sparse regions for flood prediction~\citep{gang2021improving}. In applications with limited observational data, transductive transfer learning approaches such as unsupervised domain adaptation demonstrate improved streamflow prediction~\citep{zhou2022flooddan} within local regions. Despite this, we seek to address the challenges of data sparsity in \textit{global} regions by leveraging information from data-rich areas to induce transfer learning.

This paper presents an attention-based domain adaptation streamflow forecaster targeting data-sparse regions. Our approach leverages a data-rich source domain to enable suitable forecasts in a data-constrained target domain by employing a sequence-to-sequence model for domain adaptation. Herein, we apply an attention mechanism to two separate, or \textit{private}, encoder-decoder recurrent neural networks (RNN) in both the source and target domains. This simultaneously generates future streamflow sequences and distinguish between the source and target features using adversarial learning. We demonstrate the effectiveness of our approach on data from the US (source) and Chile (target) regions extracted from Catchment Attributes and Meteorology for Large-sample Studies datasets~\citep{newman2015development, alvarez2018camels}. Our approach outperforms baseline transfer learning techniques on datasets with limited streamflow observations.

\begin{figure*}[ht!]
  \vspace{-1em}
  \centering
  \includegraphics[width=0.9\linewidth]{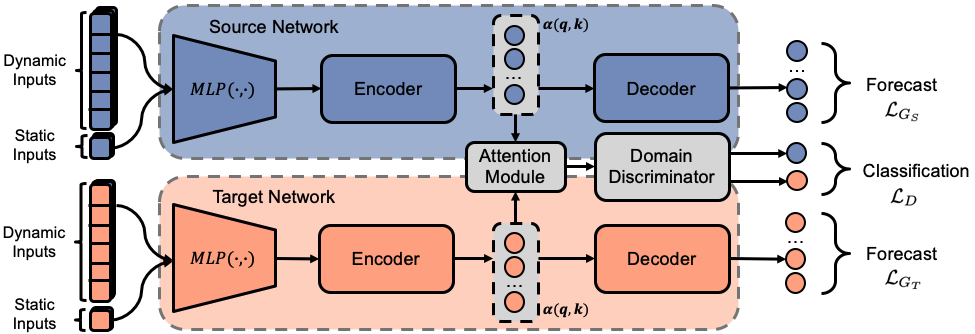}
  	  \caption{Main architecture of the attention-based domain adaptation forecaster (best seen in color). An encoder-decoder with attention is trained privately for the source and target domains for sequence generation, and the domain discriminator classifies the origin of the extracted features.} 
  \label{fig:main_arch}
\end{figure*}

\section{Methods}
\subsection{Overall Process}
In this section, we detail our attention-based domain adaptation approach for effective streamflow forecasts in data-sparse regions. We built two separate, or \textit{private}, sequence-to-sequence networks with an attention module for the source and target domains, and trained adversarially to refine domain alignment and prediction. Figure~\ref{fig:main_arch} shows the end-to-end process of the model that aims to leverage rich data samples from the source domain, denoted as $\LagrD_S$, to induce knowledge transfer on limited samples from a target domain, denoted as $\LagrD_T$. We train a source sequence generator $G_S$ and a target sequence generator $G_T$ of streamflow predictions. In the context of adversarial transfer learning, we develop a discriminator function $D$ that discerns between the domain origins through a binary classification method. We establish our optimization problem as a minimax training objective shown as:

\begin{equation}
\underset{G_S, G_T}{min}\underset{D}{max}~~\Lagr_{G_S}(\LagrD_S;\theta_S) + \Lagr_{G_T}(\LagrD_T;\theta_T) -\lambda \Lagr_{D} (\LagrD_S, \LagrD_T; D, \theta_S, \theta_T),
\label{eq1}
\end{equation}

where $\theta_S$ and $\theta_T$ are the parameters of the source and target network, respectively, and $\lambda$ is a trade off parameter to balance the two objectives. We detail the methods of our approach in the following.

\subsection{Input to Encoder Network}
The prediction of streamflow can be determined by the hydrological features of a river basin that are independent of one another~\citep{addor2017camels}. Herein, we define \textit{dynamic inputs} as the features of water catchments that are tracked throughout time (e.g., min and max air temperature, precipitation, vapor pressure) and \textit{static inputs} as the catchment attributes that remain fixed (e.g., soil type, climatological variables). In order to find a suitable input representation for our domain adaptation scheme, we pass both dynamic and static inputs through a multi-layer perceptron $MLP(\cdot,\cdot)$, where the two terms are the inputs and the parameters of the model. The MLP model creates a combined hidden representation, or embedding, for both dynamic and static inputs by projecting them into a common latent space. This output representation can be fed into the encoder model.

We employ an encoder network that is privately built for the source and target domain. Given a time series of a historical input sequence of length $N$ and a set of future target sequences $\tau$, the encoder extracts the embedded hydrological features from the historical input sequence for the first $N$ time steps in each domain. We built the encoder using an RNN, specifically a long-short term memory (LSTM) network~\citep{hochreiter1997long}, that takes an input of $X=[x_n]_{n=1}^{N}$, where each input $x_n$ is the embedded dynamic and static inputs generated by the MLP at time step $n$. The encoder extracts features through the recurring cell states to create a set of hidden representations. The output of the encoder is fed into an attention layer, where attention vectors are computed with the decoder to make $N+\tau$ future predictions.

\subsection{Attention-Decoder Network \label{sec:attn-dec}}
Attention is computed separately for the source and target networks and generates a similarity score $score(h_n, \bar{h}_s)$, where $h_n$ is the current state of the decoder at time $n$ and $\bar{h}_s$ is the set of hidden states of the encoder. A softmax function is then applied to normalize the similarities into probabilities to obtain our attention weights for each network, denoted as $\alpha_{n,s}$. A context vector $c_n$ at time $n$ is then computed as the weighted sum of the encoder hidden states. In the case of a sequence-to-sequence model, we used an additive attention score from the work in~\citep{bahdanau2014neural}. The decoder network takes the context vector for each private network to produce the set of $N+\tau$ streamflow predictions over the target sequence for the source and target domains. In addition, we employ teacher forcing to enable the decoder to generate a set of future predictions $\hat{Y} = [\hat{y}]_{n=N+1}^{N+\tau}$ over $\tau$ time steps.

To compute the model's training performance for the source loss $\Lagr_{G_S}$ and target loss $\Lagr_{G_T}$ in generating a streamflow prediction, we use the Nash-Sutcliffe efficiency (NSE) \citep{nash1970river}, which is a statistical measurement that predicts the skill of a hydrological model. NSE is defined in Equation~\ref{eq2} as:

\begin{equation}
NSE = 1 - \frac{ \sum_{n=1}^{N} (Q_{m,n} - Q_{o,n})^2}{ \sum_{n=1}^{N} (Q_{o,n} - \bar{Q}_{o})^2},
\label{eq2}
\end{equation}

where $Q_{m, n}$ is the predicted streamflow generated by model $m$ at time step $n$, $Q_{o,n}$ is the observed streamflow $o$ at time step \(n\), and $\bar{Q}_o$ is the mean observed streamflow. The value of the NSE varies on the interval $(-\infty, 1]$, where $NSE = 1$ is desirable. 



\subsection{Domain Discriminator}
In order to mitigate the domain shift between two global regions, we employ a domain adaptation technique using adversarial transfer learning to induce better distribution alignment between the source and target domain. We first map the context vectors $c_{S, n}$ and $c_{T, n}$ of the source and target networks at time step $n$, respectively, into a shared latent space to produce a feature vector $h_c$. We then employ a discriminator to identify the origins of the two domains. The domain discriminator is a binary classifier $D = MLP(h_c; \theta_D)$, where $\theta_D$ are the parameters of the discriminator function. The discriminator is trained to optimize the maximum loss in classifying between the two domains. The loss of the discriminator $\Lagr_{D}$ uses binary cross-entropy as its objective function.

Adversarial training balances the training objectives of the sequence generators and the domain discriminator. We use the NSE objective function to minimize the loss of sequence generators $\Lagr_{G_S}$ and $\Lagr_{G_T}$, and binary cross-entropy to maximize the loss for the domain discriminator $\Lagr_D$, as shown in Equation~\ref{eq1}. In other words, while $D$ tries to classify between the source and target domain using the attention module, $G_S$ and $G_T$ tries to confuse $D$ by producing features that are indistinguishable in the shared latent space. We employ a coefficient $\lambda$ multiplied to the discriminator loss $\Lagr_D$ in order to balance the trade-off between the two objectives. In our experiments we set the coefficient $\lambda$ to be 0.1.

\section{Experiments and Results}
We collected data from the CAMELS-US~\citep{newman2015development} and CAMELS-Chile (CAMELS-CL)~\citep{alvarez2018camels} as our source and target domains, respectively. The CAMELS-US is a large sample benchmark dataset that has covered temporal and geospatial information related hydrology across 671 basins in the United States, 531 of which were used for this experiment. The CAMELS-CL is also a sufficiently large benchmark dataset with 516 basins covering the country of Chile. To show the data sparsity, we downsized the CAMELS-CL dataset to 253 basins in the Chile region with approximately 10\% of missing streamflow data. 
We split the dataset into a train set from October 1st, 1999 to September 30th, 2000, a validation set from October 1st, 1988 to September 30th, 1989, and a test set from October 1st, 1989 to September 30th, 1999. 


We test our attention-based domain adaptation forecasting approach against baseline models on 24hr lead time streamflow prediction. Specifically, we employ an LSTM and GRU model from the open-source work of~\cite{kratzert2022joss} for network transfer learning by leveraging a pre-trained source model to induce transfer learning in the target domain. The hyperparameters of each model in this experiment consisted of 128 hidden units and 1 output regression layer with a dropout of 0.4 used for forecasting daily streamflow. We initialize the learning rate \( \eta \) for each model to 0.001 on the first epoch and adjust to 0.0005 over the remaining epochs. In addition, we use ADAM as the optimizer.

\begin{table}[ht]
\centering
\caption{Streamflow metrics that measure the performance of hydrological models. We detail each metric with original references, equations, and the range of scores with desirable values.}
\begin{tabular}{l l l}
\hline
Metrics & Equation & Range \\ 
\hline
\begin{tabular}[c]{@{}l@{}} Nash Sutcliffe Efficiency \\ \citep{nash1970river} \end{tabular} 
& $1 - \dfrac{\sum_{n=1}^{N} (Q_{m,n} - Q_{o,n})^2}{ \sum_{n=1}^{N} (Q_{o,n} - \bar{Q}_{o})^2}$ 
& \begin{tabular}[c]{@{}l@{}} $(-\infty, 1]$; values closer to \\ 1 are desirable. \end{tabular} \\
\hline
\begin{tabular}[c]{@{}l@{}} Kling-Gupta Efficiency \\ \citep{gupta2009decomposition} \end{tabular}
& $\begin{array}[c]{@{}l@{}} 1 - \sqrt{(\mathit{r}-1)^{2} + (\alpha-1)^{2} + (\beta-1)^{2}} \end{array}$
& \begin{tabular}[c]{@{}l@{}} $(-\infty, 1]$; values closer to \\ 1 are desirable. \end{tabular} \\ 
\hline
\begin{tabular}[c]{@{}l@{}} $\alpha$-NSE decomposition \\ \citep{gupta2009decomposition} \end{tabular}
& $\sigma_{m} / \sigma_{o}$ 
& \begin{tabular}[c]{@{}l@{}} $(0, \infty)$; values closer to \\ 1 are desirable. \end{tabular} \\
\hline
\begin{tabular}[c]{@{}l@{}} $\beta$-NSE decomposition \\ \citep{gupta2009decomposition} \end{tabular}
& $(\mu_{m} - \mu_{o}) / \sigma_{o}$ 
& \begin{tabular}[c]{@{}l@{}} $(-\infty, \infty)$; values closer to \\ 0 are desirable. \end{tabular} \\
\hline
\end{tabular}
\label{tab:table1}
\end{table}

Table~\ref{tab:table1} describes the streamflow metrics to measure the performance of the aforementioned hydrological models. NSE~\cite{nash1970river} is used as the primary metric for each experiment (see Section~\ref{sec:attn-dec}). Kling-Gupta Efficiency (KGE)~\citep{gupta2009decomposition} is the square root of squared sums of the linear correlation between observations and simulations $\mathit{r}$, the measurement of the flow variability error $\alpha$, and the bias term $\beta$. Metric $\alpha$-NSE~\citep{gupta2009decomposition} is the fraction of the standard deviations of predicted streamflow and observations. $\beta$-NSE~\citep{gupta2009decomposition} is the difference of the mean predicted streamflow and mean observation divided by the standard deviation of the observations. In addition, we include the number of basins where NSE $<$ 0.

\begin{table*}[ht]
\centering
\caption{Model comparisons between our proposed method and baseline models using the hydrological streamflow measurements. The statistics were averaged across 5 runs.}
\begin{tabular}{l c c c c c}
    \hline
    \multicolumn{6}{c}{CAMELS-US to CAMELS-CL} \\
    \hline
    Model & NSE & KGE & $\alpha$-NSE & $\beta$-NSE & NSE $\mathrm{ < 0}$ \\
    \hline
    LSTM-TL & $0.32 \pm 0.01$ & $0.31 \pm 0.01$ & $0.59 \pm 0.01$ & $-0.10 \pm 0.01$ & $66.40 \pm 1.86$ \\
    \hline
    GRU-TL & $0.11 \pm 0.02$ & $0.08 \pm  0.02$ & $0.43 \pm 0.03$ & $-0.17 \pm 0.01$ & $92.60 \pm 5.00$ \\
    \hline
    Our approach & $\textbf{0.37} \boldsymbol{\pm} \textbf{0.02}$ & $\textbf{0.41} \boldsymbol{\pm} \textbf{0.03}$ & $\textbf{0.68} \boldsymbol{\pm} \textbf{0.03}$ & $\textbf{–0.06} \boldsymbol{\pm} \textbf{0.03}$ & $\textbf{63.80} \boldsymbol{\pm} \textbf{1.60}$ \\
    \hline     
\end{tabular}
\label{tab:table2}
\end{table*}

\begin{figure}[ht]
  \centering
  \vspace{-1.0em}
  \includegraphics[width=0.7\linewidth]{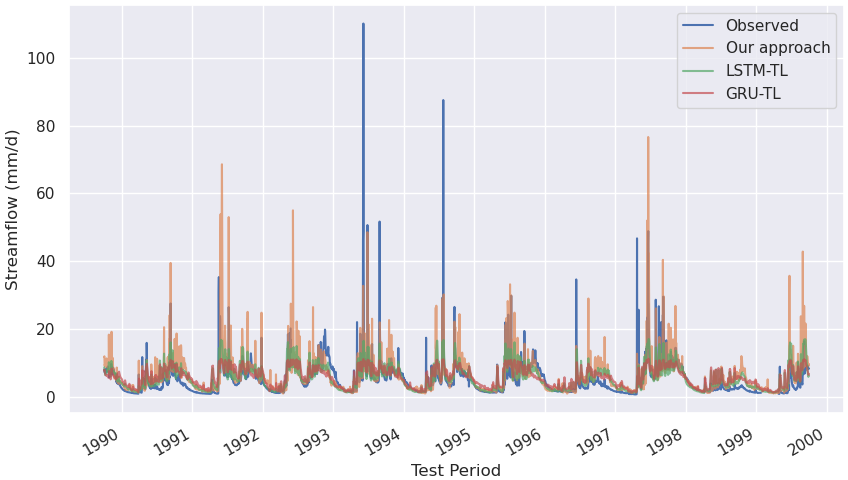}
  	  \caption{Comparisons on the prediction of streamflow between our attention-based domain adaptation streamflow forecaster and the RNN baselines LSTM-TL and GRU-TL. The x-axis is the test period from October 1st, 1989 to September 30th 1999 and the y-axis is streamflow values in mm/d.} 
  \label{fig:compare_predictions}
\end{figure}

Table~\ref{tab:table2} reports the median scores for all metrics except $NSE \mathrm{ < 0}$. The performance results for each model on 24hr lead-time streamflow prediction were averaged across 5 runs. Our attention-based domain adaptation approach shows greater performance compared to the baseline models across all hydrological metrics used in this experiment. Specifically, our approach outperformed both the network-based transfer learning RNNs in terms of median NSE and median KGE, showing 0.05-0.16 NSE and 0.09-0.33 KGE score improvements. In terms of $\alpha$-NSE and $\beta$-NSE, our approach showed a significant increase of performance compared to the two RNN transfer learning baselines at very small deviations. Lastly, we observed the number of basins with subpar NSE scores (i.e., $NSE \mathrm{ < 0})$: our approach reported approximately 4\% fewer water basins than LSTM-TL and approximately 31\% fewer basins than GRU-TL. In addition, we further demonstrate how the results of our median NSE scores more closely match the observed streamflow compared to the baselines in Figure~\ref{fig:compare_predictions}. We note that the results are particularly impressive since we only used one year of training data rather than the full ten year dataset.

\begin{figure}
  \centering
  \includegraphics[width=0.7\linewidth]{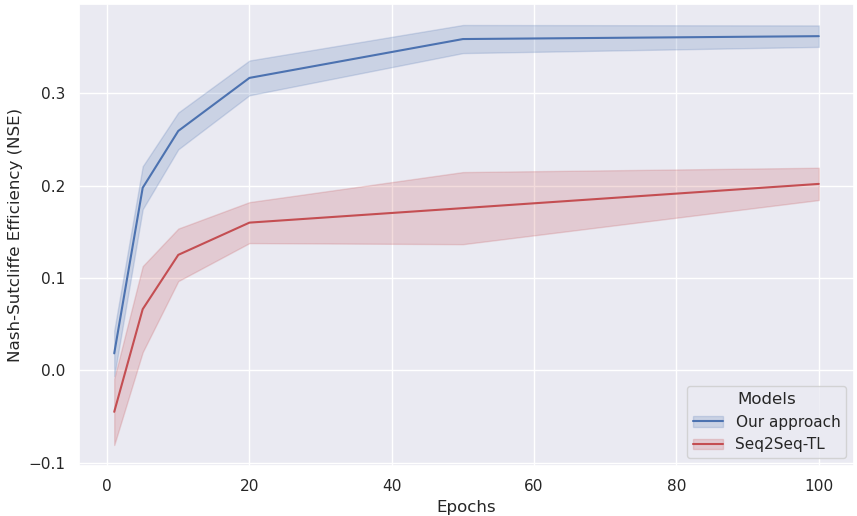}
  	  \caption{Ablation study comparing the streamflow performance between our attention-based domain adaptation method using adversarial training and a sequence-to-sequence model with only attention viz., \textit{Seq2Seq-TL}. The results are trained over 100 epochs and averaged across 5 runs.} 
  \label{fig:ablation}
\end{figure}

We report an ablation study that compares our attention-based domain adaptation streamflow forecaster with a sequence-to-sequence transfer learning model using only attention viz., \textit{Seq2Seq-TL}. We implemented the Seq2Seq-TL model by pre-training on the source dataset and performing network-based transfer learning to the target dataset. We replicated the same hyperparameters in Seq2Seq-TL from our approach and compared the performance over 100 training epochs. From Figure~\ref{fig:ablation}, the plots clearly display the decrease in streamflow prediction in Seq2Seq-TL when adversarial training is not used. Hence, this verifies the robustness of our approach in inducing domain invariance for better feature alignment between the source and target domains.

\section{Impact on Climate Change and Conclusion}
Streamflow forecasts are critical to guide water resource management in a changing climate. Climate change is expected to alter the timing and amount of precipitation, which in turn affects the timing and magnitude of streamflow. Therefore, accurate streamflow forecasts become even more critical as we face more frequent and severe weather events such as droughts and floods due to climate change. In many regions, access to clean water is often a challenge, and water resources management is critical for human health, food security, and economic development. Climate change may exacerbate existing water challenges and make it even harder to manage water resources sustainably. Climate-smart, accurate streamflow forecasts can be critical in these regions, as they can help to improve water resource management and support adaptation to changing conditions.

In this paper, we propose a domain adaptation forecaster via attention for streamflow in data-sparse regions. In our approach, we built an encoder-decoder deep learning framework where attention is applied to capture salient hydrological features across water basins. We apply a domain discriminator to discern between the origin of the source and target features through adversarial learning. The results on the CAMELS-CL region with approximately 10\% missing streamflow data show that, on average, our model generalizes better ($0.37$ NSE) compared to the transfer learning LSTM ($0.32$ NSE) and GRU ($0.11$ NSE) models.



\bibliography{main}
\bibliographystyle{iclr}


\end{document}